\documentclass[conference]{IEEEtran}
\usepackage{graphics}      
\usepackage[T1]{fontenc}   
\usepackage[utf8]{inputenc}
\usepackage{mathptmx}
\usepackage[pdflang={en-US},pdftex]{hyperref}
\usepackage{color}
\usepackage{booktabs}
\usepackage{textcomp}
\usepackage{float}
\usepackage{enumitem}
\usepackage{cite}
\usepackage{newtxmath}

\usepackage{algorithmic}
\usepackage{graphicx}
\usepackage{textcomp}
\usepackage{xcolor}
\def\BibTeX{{\rm B\kern-.05em{\sc i\kern-.025em b}\kern-.08em
    T\kern-.1667em\lower.7ex\hbox{E}\kern-.125emX}}
\begin{document}

\title{Adversarial Attacks against Neural Networks in Audio Domain: Exploiting Principal Components \\
}

\author{
\IEEEauthorblockN{Ken Alparslan\textsection}
\IEEEauthorblockA{\textit{Department of Computer Science} \\
\textit{Conestoga College}\\
Waterloo, ON, CA \\
kalparslan6724@conestogac.on.ca}
\and
\IEEEauthorblockN{Yigit Alparslan\textsection}
\IEEEauthorblockA{\textit{Department of Computer Science} \\
\textit{Drexel University}\\
Philadelphia, PA, US \\
ya332@drexel.edu}
\and
\IEEEauthorblockN{Dr. Matthew Burlick}
\IEEEauthorblockA{\textit{Department of Computer Science} \\
\textit{Drexel University}\\
Philadelphia, PA, US \\
mjb528@drexel.edu}
}
\maketitle
\begingroup\renewcommand\thefootnote{\textsection}
\footnotetext{These co-first authors contributed equally.}
\endgroup

\begin{abstract}
Adversarial attacks are inputs that are similar to original inputs
but altered on purpose. Speech-to-text neural
networks that are widely used today are prone to misclassify
adversarial attacks [1], [2], [3]. In this study, first, we investigate
the presence of targeted adversarial attacks [4], [11] by altering
wave forms from Common Voice data set. We craft adversarial
wave forms via Connectionist Temporal Classification Loss
Function [13], [14], [15] and attack DeepSpeech [12]– speech-to-text
neural network implemented by Mozilla. We achieve
100\% adversarial success rate (zero successful classification by
DeepSpeech) on all 25 adversarial wave forms that we crafted.
Second, we investigate the use of PCA as a defense mechanism
against adversarial attacks.We reduce dimensionality by applying PCA to these 25
attacks that we created and test them against DeepSpeech.
We observe zero successful classification by DeepSpeech, which suggests
PCA is not a good defense mechanism in audio domain. Finally,
instead of using PCA as a defense mechanism, we use
PCA  this time to craft adversarial inputs under a black-box
setting with minimal adversarial knowledge. With no knowledge
regarding the model, parameters, or weights, we craft
adversarial attacks by applying PCA to samples
from Common Voice data set, and achieve 100\% adversarial
success under black-box setting again when tested against
DeepSpeech. We also experiment with different percentage
of components necessary to result in a classification during
attacking process. In all cases, adversary becomes successful.
\end{abstract}


\section{Introduction}
Numerous recent studies have demonstrated how Deep Neural
Network (DNN) classifiers can be fooled by adversarial examples [1], [2], [3], [4], [5] in which an attacker adds perturbations
to an original sample, causing the classifier to misclassify the
sample [6]. Adversarial attacks that render DNNs vulnerable
in real life represent a serious threat, given the consequences of
improperly functioning autonomous vehicles, malware filters,
bio-metric authentication systems or surveillance systems [7], [8], [9]. Social network companies, crime-fighting organizations, law enforcement, and various commercial interests
may wish to be able to use image classification and speech-to-
text transcription tools with defensive mechanisms that are
robust to adversarial attacks [2], [11], [29] in order to reduce these attacks' effectiveness. Wide usage of DNNs makes the problem
of creating robust and secure DNNs even more important in
safety-critical applications.
We are interested in exploring whether there are any particular
characteristics of audio recognition which seem to make adversarial
attacks successful in this area. Current
attacks that have been studied were primarily in the image
domain [4], [5], [6], [7], such as those of Carlini [2] and Papernot
et al [17].  Such attacks have been studied as proof-of-concepts,
where adversarial attackers are assumed to have full knowledge
of the classifier (e.g. model, architecture, model weights,
parameters, training and testing data sets). The strongest attack
in the literature at the time of writing this article is Carlini’s
attack [11] based on the $L_{2}$ norm, and it is a white-box attack
requiring full knowledge of the model. Much of this
research has been interested in developing the most effective
attacks possible, to be used as standards against which to test
the robustness of classifier RNNs [2]. With less knowledge of
the classifier model, the effectiveness of the attack decreases.
There is also interest in crafting attacks that assume minimal
knowledge of the adversary –that is, attacks under black-box
setting – regarding the classifier model, since in most real world
applications the adversary does not have access to the
classifier’s parameters unless the adversary is an insider. \\
We are inspired by work on image domain [2], [3], [4], [5], [6] and
investigate the audio domain in this paper.
We use Common Voice Data set, which is widely used as a
standard audio data set in the literature [11]. Being a widely used
data set, it provides a common ground for different research
labs all around the world. We use 25 samples from Common
Voice to create our adversarial attacks. We also use Deep-
Speech, a state-of-the-art speech-to-text transcription neural
network implemented by Mozilla [12] to test our attacks. In the
literature, one can see that Recurrent Neural Networks are often
used for audio transcription/decoding systems [13], [14], [15].
RNNs are effective to map an audio signal to a sequence of
probability distributions over individual characters. DeepSpeech, for example, is a recurrent neural network that uses Long Short-Term Memory (LSTM) Neural Network Architectures. Connectionist Temporal Classification (CTC) [13] is a method of training a sequence-to-sequence neural network
when the alignment between the input and output sequences
is not known. DeepSpeech uses CTC because the inputs are
an audio sample of a person speaking, and the unaligned transcribed
sentences, where the exact position of each word in
the audio sample is not known.
In order to study neural networks in the audio recognition
field, we focus on creating adversarial attacks that are crafted
under white-box setting, as well as black box setting. For
white-box approached attacks [11], we represent audio as a
N-dimensional vector x. Each element $x_i$ is a signed 16-bit
value sampled at 16KHz. Targeted attacks are crafted as
follows: One second of given waveform is split into 50 frames.
Input domain consists of single frames and the output domain
consists of characters in the a-z space. RNN (in our case,
we have a hook to the DeepSpeech's language decoder, and
RNN) takes a sequence of N frames and returns a probability
distribution over the output domain for each frame.
While CTC maps every frame to a probability distribution over
the characters, this does not directly give a probability distribution
over all phrases. The probability of a phrase is defined as
a function of the probability of each character. For in-depth analysis, we refer the reader to [25]. Carlini et al [11] gives two short definitions. A \textit{sequence} $\Pi$ reduces to a shortened sequence p
if starting with $\Pi$ and making the following two operations
(in order) yields p:
\begin{enumerate}
    \item Remove all sequentially duplicated tokens. 
    \item Remove all e tokens
\end{enumerate}

For example, the sequence a a b e e b reduces to a b b.
$\Pi$ is an alignment of p with respect to target labels y if
\begin{enumerate}
    \item $\Pi$ reduces to p,
    \item the length of p is equal
to the length of y.
\end{enumerate} Carlini et al [11] defines the probability of alignment p under y as
the product of the likelihoods of each of its elements:

\begin{equation}
\centering
  Pr(\Pi|y) = \prod_{i}^{} y^{i}_{\pi^{i}}
\end{equation}
The probability of a given phrase p under the distribution y = f(x) becomes 
\begin{equation}
\centering
Pr(p|y) = \sum_{\pi\in\prod
(p,y)}^{} Pr(\pi|y)  = \sum_{\pi\in\prod (p,y)^{}}\prod_{i}^{} y^{i}_{\pi^{i}}
\end{equation}

As is usually done, the loss function used to train the network is the negative log likelihood of the desired phrase:
\begin{equation}
\centering
CTC-Loss(f(x), p) = -log Pr(p|f(x))
\end{equation}

Carlini et al [11] gives the following definition to decode a vector y to a phrase p, such that it best aligns to y.
\begin{equation}
\centering
C(x) = arg max_{\pi}Pr(p|f(x))
\end{equation}

The search space is non-trivial. Current literature suggests some optimization algorithm and use of dynamic programming to help find the phrase p [11]. In our study, it is approximated via Beam Search Decoding. Beam Search Decoding simultaneously evaluates the likelihood of multiple alignments $\pi$ and then chooses the most likely phrase p under these alignments. We refer the reader to [13] for a complete algorithm description.

We were able to craft adversarial attacks against Common Voice data set which achieved 100\% misclassification rate against the classifier when tested against DeepSpeech. The perturbed audio files produced by this attack, while altered, were still clearly identifiable to a human. Even visually, the wave forms are nearly indistinguishable as it can be seen in Figure \ref{fig:figure2} and \ref{fig:figure3}

\section{Related Work}
Previously, adversarial examples have focused largely on domain of images, and face detection as it can be seen in [26], [27] and [28]. In the discrete domain, text classification is also studied [19]. However, adversarial attacks in audio recognition domain remains relatively new.

\begin{enumerate}
    \item Goodfellow et al, (2013) highlighted adversarial examples as a ubiquitous threat to different neural network models [10]
    \item Papernot et al, (2016) showed the limitations of deep neural networks in image recognition [20]
    \item Adrian et al, (2018) showed that adversarial training can be used as a defense against attacks in image domain [21]
    \item Carlini et al, (2018) crafted targeted audio adversarial attacks [11]
    \item Yakura et al, (2019) demonstrated the possibility of an over-the-air audio attacks [22].

\end{enumerate}
We are inspired by the previous work, where adversarial were attacking image classifiers in [1], [2], [3], [4], [5], [6], [7]. By investigating adversarial attacks in audio domain, and looking at different ways of crafting attacks, we hope our research may help to advance the study of adversarial attacks on RNNs and defensive mechanisms to counteract them, particularly in the audio recognition domain.

\section{Problem Statement}
In this study, we are tacking the following problems:
\begin{enumerate}[label=(\roman*)]
    \item How can we produce strong adversarial examples in audio domain, i.e., transferable adversarial examples that fool a model with high confidence while requiring only a small perturbation?  
    \item How can we train a model to be robust so that there are no audio adversarial examples, or at least so that an adversary cannot find them easily?

\end{enumerate}

To answer these questions, we investigate creating audio adversarial attacks, and explore evaluating robustness of a neural network on following aspects:
\begin{enumerate}
    \item Presence/Robustness of white-box and black-box adversarial attacks (Can we embed a sentence into a song? Can we force DeepSpeech to transcribe a sentence to some other sentence that we want? Can we reduce number of frequency components so that difference is subtle enough for a human to not recognize, but big enough for DeepSpeech to transcribe it poorly without any adversarial knowledge regarding model weight, parameters, and architecture?)
    \item Use of PCA to craft adversarial attacks under black-box setting with minimal adversary knowledge regarding the model (Can we use PCA to alter the audio and result in misclassification by the RNN?)
    \item Use of PCA as a defense mechanism (Can we use PCA to smooth out the adversarial noise, and recover the original content?)
\end{enumerate}

We have the following basic approach in this study.
\begin{enumerate}
    \item Use PCA to craft adversarial attacks in audio under black-box setting with minimal adversarial knowledge
    \item Use CTC loss function, gradient descent, on a Long short-term memory architecture (LSTM) to craft adversarial attacks in audio under white-box setting with full adversarial knowledge.
    \item Apply attacks to DeepSpeech and compare results.
    \item Use PCA as a defense mechanism to smooth out the adversarial noise in order to see if robustness is improved against adversarial attacks
\end{enumerate}

When we are creating \emph{targeted} attacks, we assume that the adversary has complete access to a neural network, including the architecture and all
parameters, and can use this in a white-box manner. This is a conservative assumption. There have been various attempts [15], [2], [20], [42], [39] at constructing defenses that increase the robustness of a neural network, defined as a measure of how easy it is to find adversarial examples that are close to their original input. To come up with the strongest attacking algorithm, full adversarial knowledge was assumed in most of the cases. However, in real world, one might find that it is not the case most of the time. A malicious input might not know for instance, how Amazon Alexa was trained to pass some evil hidden voice commands. Another example might be ambient noise added to the input. In the ambient noise case, the input is not altered on purpose. The input happens to be adversarial because of the accidental noise in the environment/channel. In that case, the noise should not result in misclassification if the underlying RNN is robust against noise. For this reason, we also study adversarial attacks that are crafted under black box setting with minimal adversarial knowledge. To craft those attacks, we apply PCA to reduce dimensionality of the audio signal. In our case, dimensions are frequency components that makes up the audio signal after taking the Fast Fourier Transform (FFT). We usually pick a high sampling rate for the FFT to avoid aliasing such as 16000 Hz. Later, we reconstruct the audio so that we only keep some percentage of the original information. (mainly 10\%,35\%,65\%, 80\%,95\%). Since input length is different for each case, we can't fix the number of components that are used to reconstruct the original signal. Instead we fix the percentage of reconstructed information (that is, how much of the content is retained after reconstruction). For 25 of the samples from Common Voice data set, we apply PCA where we keep 10\%, 35\%, 60\%,90\% of the components after reconstruction. Applying PCA to an audio signal that is under 5 seconds takes about 2-3 seconds. Entire process is completed over the course of 3-4 hours.

We create adversarial attacks and investigate the differences by plotting the magnitude (loudness) with respect to time. We also plot the power with respect to frequencies. Essentially, we look for differences in time domain as well as frequency domain. We use magnitude of the components (in decibel (dB)) as a metric to keep track of how much distortion is added. We also calculate the normalized edit distance to quantify how similar two strings are(First string being the original text, second string being the output of DeepSpeech). Normalized edit distance is defined as the number of steps to convert one string to another where each action can only be 'insertion, deletion, substitution'. This helps us keep track of the success of targeted adversarial attacks when tested against DeepSpeech. 
Figure \ref{fig:figure1} shows the magnitude of one of the samples over time. When we create an adversarial attack by adding adversarial noise to it, we see that magnitude plot of the waveform has less number of spikes, and more smooth silhouette. This might be explained by the fact that we always remove frequencies from the given input whilst creating adversarial attacks. One can also think of it as 'adding silence'. For this reason, our distortion metric in dB is always negative, since the loudness of the adversarial input will be less than that of the original waveform we started with. Because the perturbation introduced is quieter than the original signal, the distortion is a negative number, where smaller values indicate quieter distortions.
\begin{equation*}
    dBx(\delta) = dB(\delta) -  dB(x)
\end{equation*}

We also use a similarity index to indicate the sequences' similarity as a float in the range [0, 1]. 
 \begin{center}
 Similarity = 2.0*M / T
 \end{center}
 where T is the total number of elements in both sequences, and M is the number of matches.
 Note that this is 1.0 if the sequences are identical, and 0.0 if they have nothing in common.

\section{Data}

We use 'Common Voice' data set to craft our adversarial examples, which is widely used as a standard audio data set in the literature. Being a common data set, it provides a ground for different research labs all around the world. We use 25 samples from Common Voice to create our adversarial attacks. We also use DeepSpeech, a state-of-the-art speech-to-text transcription neural network implemented by Mozilla to test our attacks.

\section{Results \& Observations}

Quick Summary for results are as follows:
\begin{enumerate}
    \item White-box targeted attacks result in 100\% adversarial success (DeepSpeech can fails to transcribe the original input and transcribes the targeted sentence instead). So, all 25 targeted attacks result in misclassification when tested against DeepSpeech. 
    \item We can successfully embed voice in music. DeepSpeech transcribes Bach Cello Suite No:1 as "evil" for instance
    \item We can successfully target silence, that is humans can hear the sentence, but DeepSpeech can't
    \item PCA fails to smooth out adversarial noise.
    \item PCA succeeds as a way of creating black-box attacks. We were able to reconstruct 25 samples from Common Voice Data set, by keeping 95\% of the frequency components, while achieving 100\% adversarial success rate. When we kept 10\% of the frequencies, we also achieve 100\% adversarial success rate, but the input sentence sounds very corrupted to ear.
    \item In the case of 95\% reconstruction, we achieve average similarity ratio of 87.85\% percent and normalized edit distance of 9, whereas in the case of 10\% reconstruction, we achieve average similarity ratio of 27.42\%, and normalized edit distance of 46. We encourage the user to refer to \ref{tab:table1} for intermediate PCA Reconstruction percentages. We also encourage the user to listen to crafted attacks here.
    \item We can successfully embed voice in music. This means an audio that sounds as a song to a human being might contain hidden voice commands.
    \item Targeting a long text when the input audio is short becomes very computationally expensive, whereas targeting a short text such as "evil" when the input audio is long and is rich in frequencies.
    \item Average distortion among adversarial attacks is 28.8232 dB (absolute value is taken), which is equivalent of ambient noise in a recording.
\end{enumerate}

\subsection{Black-Box Attacks}
 Black-Box attacks are not targeted. They are weaker but their advantage is that they don't have to have full adversarial knowledge regarding the model. We can use PCA as a way of crafting adversarial attacks. In figure \ref{fig:figure3}, the input file says "why should one halt on the way?". There is no targeted text. We apply PCA, and reconstruct the audio based on the 35 out of 45 frequency components (just for this sample, 95\% refer to first 35 components since input length and frequency components are never fixed in speech) so that 95\% of the information is stored. DeepSpeech's output, which is defined as the transcribed text, is altered but still very close to the original. DeepSpeech's output string has a Normalized Edit Distance of 9 to the original, which means it would take 9-unit step actions to convert one string to another. Original audio and the adversarial audio are in this case of 87.85\% similarity.

 In figure \ref{fig:figure4}, the input file says "why should one halt on the way?". There is no targeted text. We apply PCA, and reconstruct the audio based on the 35 out of 45 frequency components so that 10\% of the information is stored. transcribed text is very much altered, and DeepSpeech performs very poorly, and fails to recognize it. DeepSpeech's output string has a Normalized Edit Distance of 46, which means it would take 46-unit step actions to convert one string to other. Original audio and the adversarial audio are in this case of 27.39\% similarity. We repeat this step for 25 samples of the Common Voice data set. We average the similarities, and normalized edit distances over 25 inputs, while experimenting with different percentages of the original content that is retained during PCA reconstruction.
 Mainly, we experiment with 10\%, 35\%, 65\%, 80\%, and 95\% of the original frequency components during reconstruction phase of PCA. We name these untargeted adversarial attacks under black-box approach '10\%, 35\%, 65\%, 80\%, and 95\% attacks' to make writing easier.
 
 \subsection{White-Box attacks}
 We notice that creating an adversarial gets much more difficult if the targeted text is long. Every extra character in the target requires an increase in, therefore for a longer text, amount of required distortion increases.  (In our case adding more adversarial noise). However, conversely, we observe that the longer the initial source phrase is, the easier it is to make it target a given transcription. We attribute this to the fact that we always remove frequencies from the original sample. Starting with an input rich in frequencies essentially takes less duration, and computational resources.
 
 Over 25 targeted adversarial attacks that were crafted had a mean distortion of 28.8232 dB, which is roughly equivalent of ambient noise. This validates the work of Nicholas Carlini et al [11].
 
 We lay out three approaches to study targeted attacks. First approach is to make the sentence get transcribed as "evil" (as a proof-of-concept that hidden evil voice commands can be embedded in voice. Also, longer targets are harder, so "evil" word works perfectly since it takes less time to minimize the CTC loss function). Second approach is to embed voice in music. We achieve that by embedding voice into Bach's Cello Suit No$^{o}$ 1. Third approach is to target silence, where we hide speech by adding
adversarial noise that causes DeepSpeech to transcribe nothing.

\begin{figure}[!ht]
\centering
  \includegraphics[width=0.9\columnwidth]{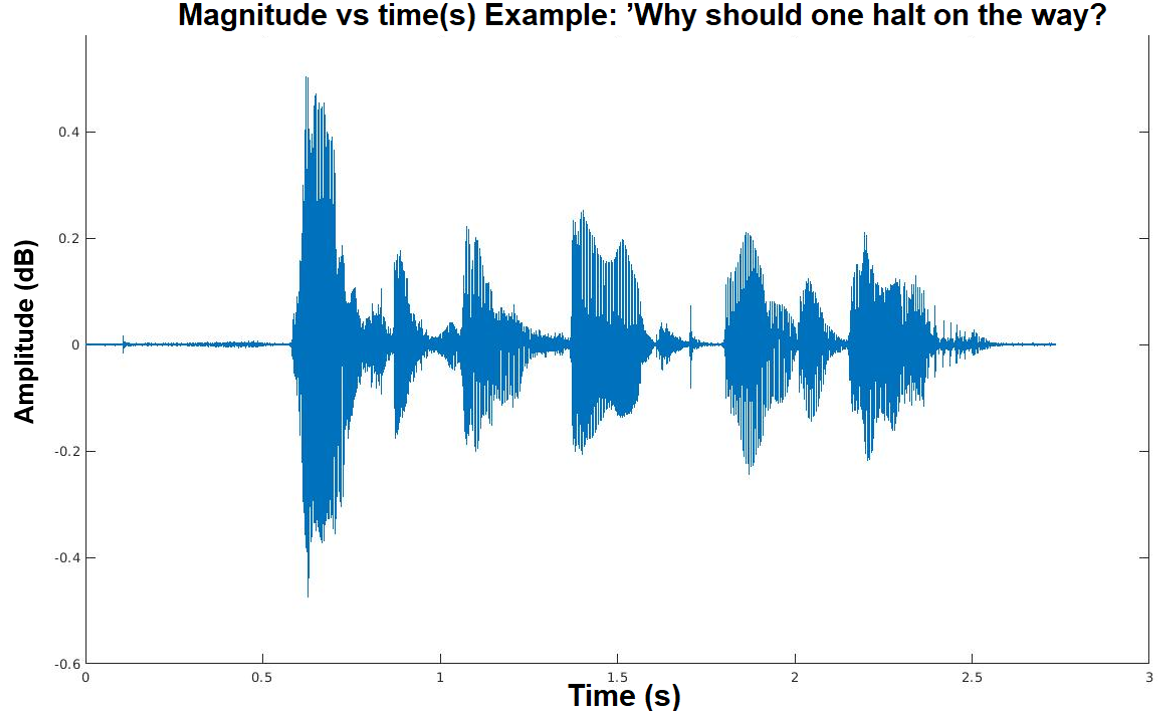}
  \caption{Magnitude vs time(s) Example: 'Why should one halt on the way?'}~\label{fig:figure1}
\end{figure}

We look for the time vs magnitude plots, time vs frequency, as well as frequency vs power plots to keep track of our perturbation. Sample plots can be seen in Figure \ref{fig:figure1},\ref{fig:figure2}, and \ref{fig:figure3}.

 \subsection{Evaluation}

\begin{figure}[!ht]
\centering
  \includegraphics[width=0.9\columnwidth]{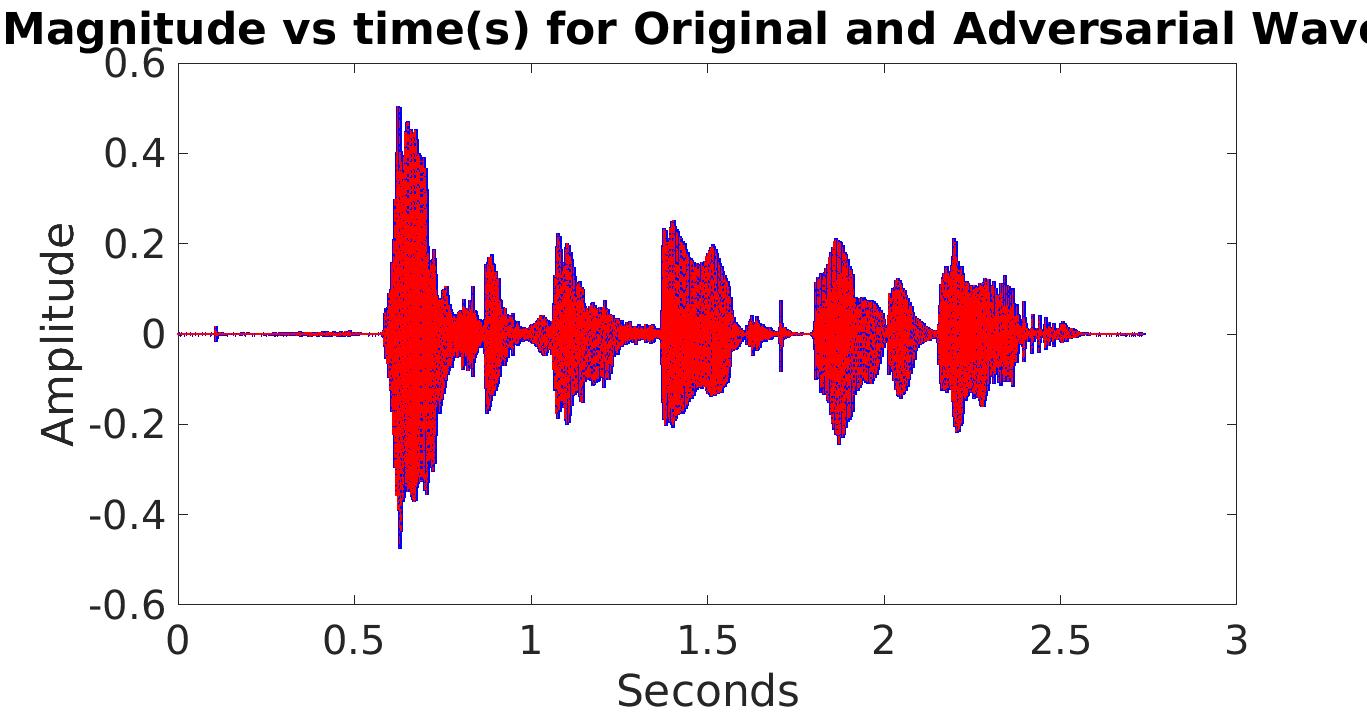}
  \caption{Magnitude vs time(s) for Original and Targeted Adversarial Waves/ When we calculate the percent MSE between two wave forms normalized with respect to the original waveform, we see 99\% similarity between wave forms.}~\label{fig:figure2}
\end{figure}
 
\begin{figure}[!ht]
\centering
  \includegraphics[width=0.9\columnwidth]{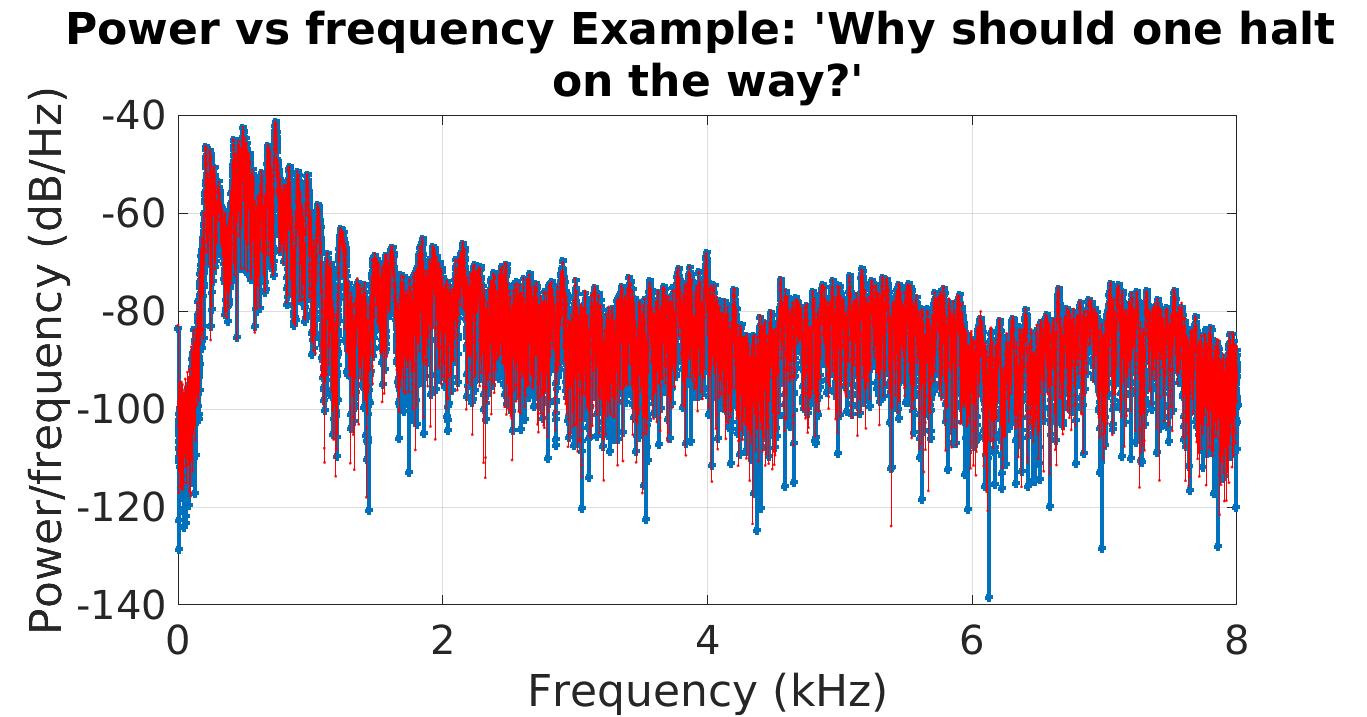}
  \caption{Power vs frequency Example: Why should one halt on the way? Red is the power spectrum of the original wave form, and blue is the power spectrum of the adversarial wave form that was crafted under white-box approach.}~\label{fig:figure3}
\end{figure}
 
As it can be seen in Figure \ref{fig:figure2} and \ref{fig:figure3} that original and adversarial wave forms are almost visually indistinguishable when plotted in time as well as frequency domain. When we crafted 25 adversarial samples taken from Common Voice Data set, we achieve 100\% adversarial success. DeepSpeech fails to transcribe the original speech in all of them. In all the cases, the targeted sentence is chosen to be "evil". Target is chosen to be relatively a short sentence (i.e only a word in this case) because it allows us to generate more samples in shorter amount of time.

\begin{figure}
\centering
  \includegraphics[width=0.9\columnwidth]{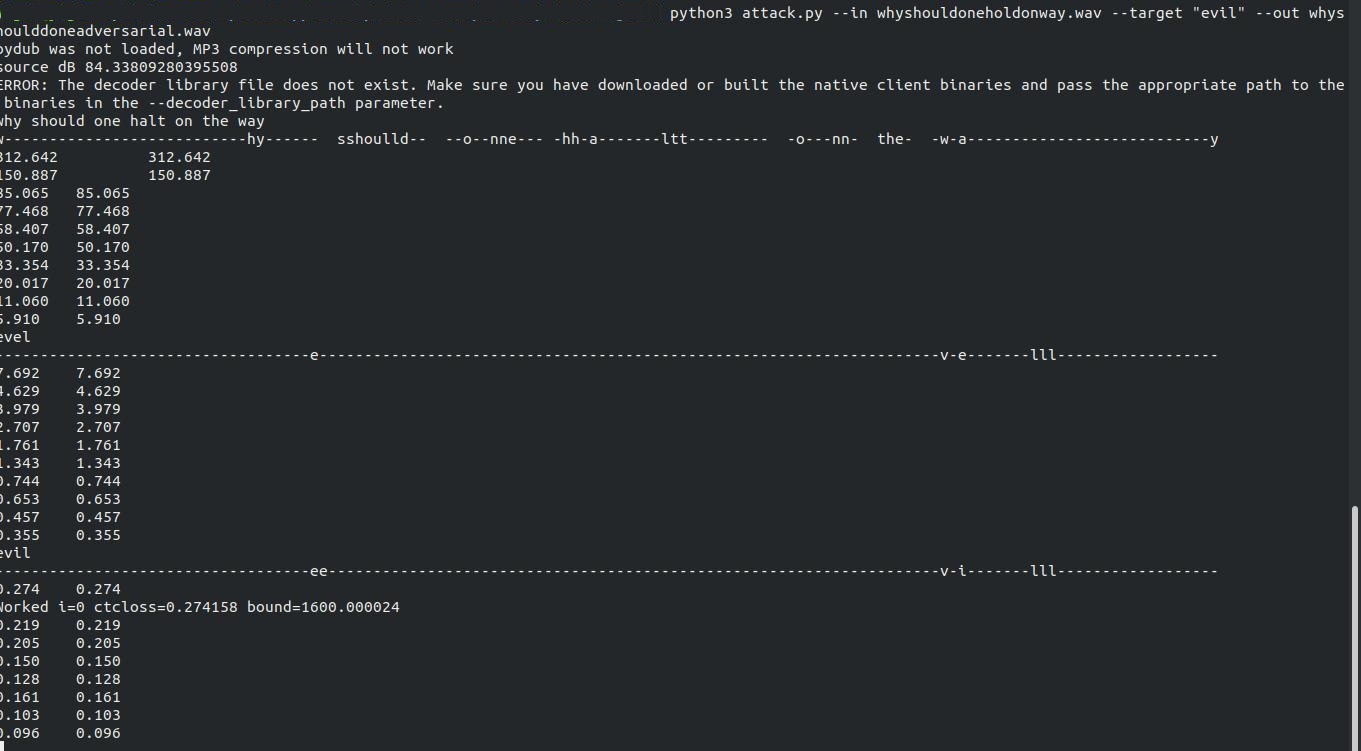}
  \caption{Example of attacking process. In this example, the input file says "why should one halt on the way?". The targeted text is "evil". }~\label{fig:figure4}
\end{figure}

\begin{figure}
\centering
  \includegraphics[width=0.9\columnwidth]{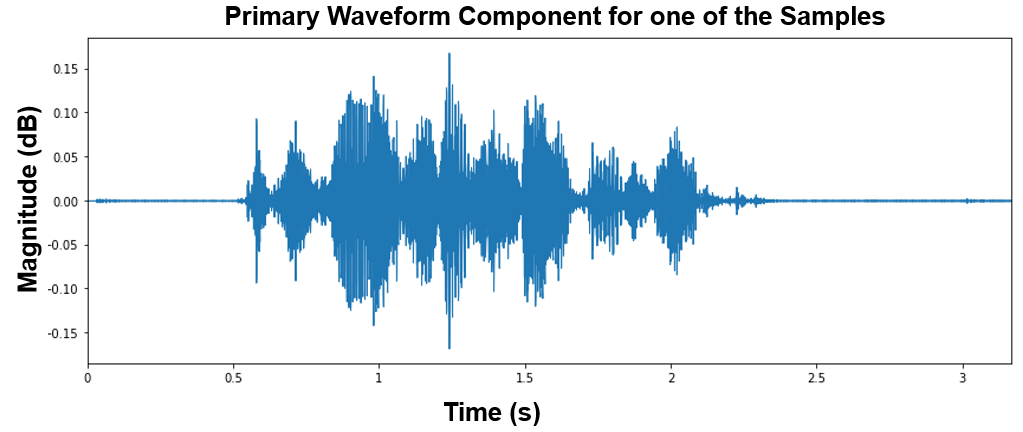}
  \caption{Primary principal component of the waveform "It seemed safe and tranquil" }~\label{fig:figure5}
\end{figure}
 
\begin{figure}
\centering
  \includegraphics[width=0.9\columnwidth]{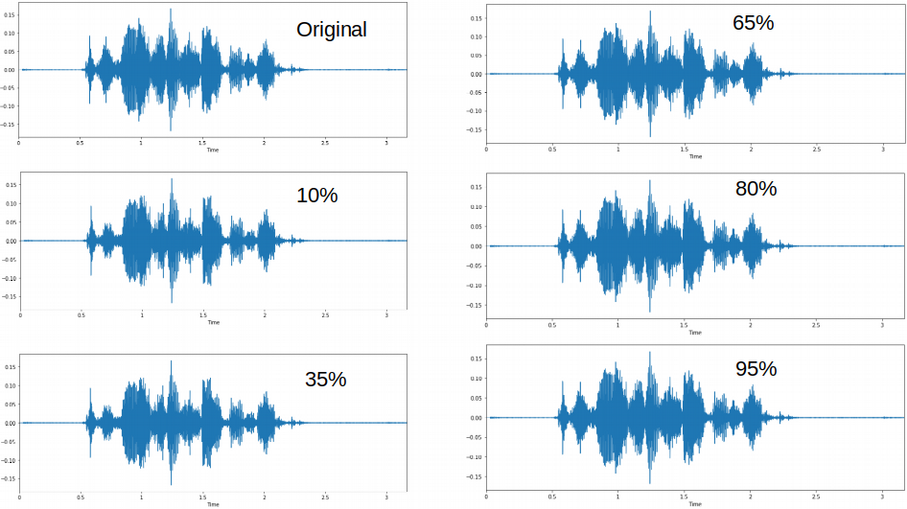}
  \caption{Comparing PCAs in frequency domain with different percentage of content retained. From top left to bottom right, we show the magnitude versus time (s) plot of the original signal as well as when PCA is applied so that we have 10\%, 35\%, 65\%, 80\%, 95\% of the original content reconstructed, respectively. }~\label{fig:figure6}
\end{figure}

 Figure \ref{fig:figure5}, and Figure \ref{fig:figure6} give insight to use of PCA as a way of crafting adversarial attack. Figure \ref{fig:figure5} shows the primary principal component of the sample "It seemed safe and tranquil." Most of the edges, and noisy spikes are eliminated compared to the original signal in Top left corner of Figure \ref{fig:figure6} with label "Original".
 
 \begin{figure}
\centering
  \includegraphics[width=0.9\columnwidth]{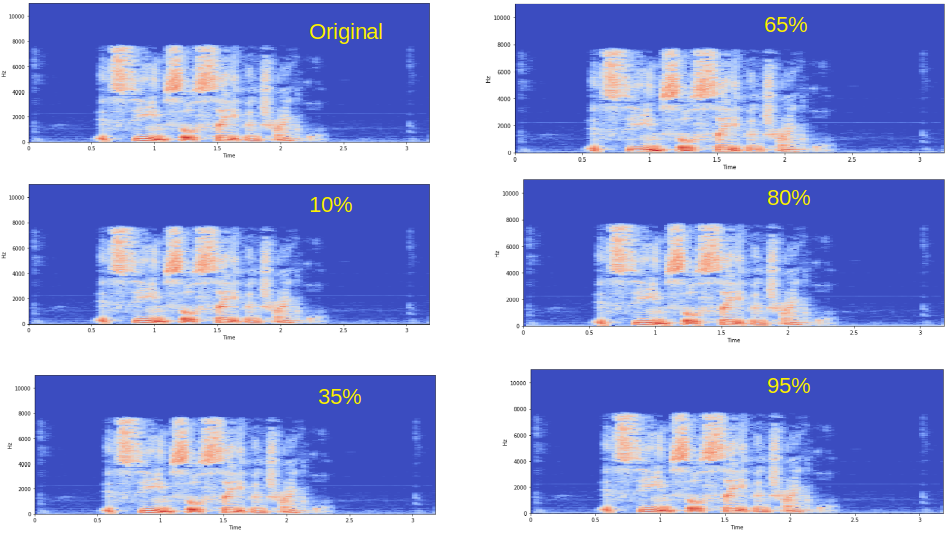}
  \caption{ Comparing PCAs in frequency domain with different percentage of content retained. From top left to bottom right, we show the frequency versus time (s) plot of the original signal as well as when PCA is applied so that we have 10\%, 35\%, 65\%, 80\%, 95\% of the original content reconstructed, respectively.  }~\label{fig:figure7}
\end{figure}

 \begin{figure}
\centering
  \includegraphics[width=1\columnwidth]{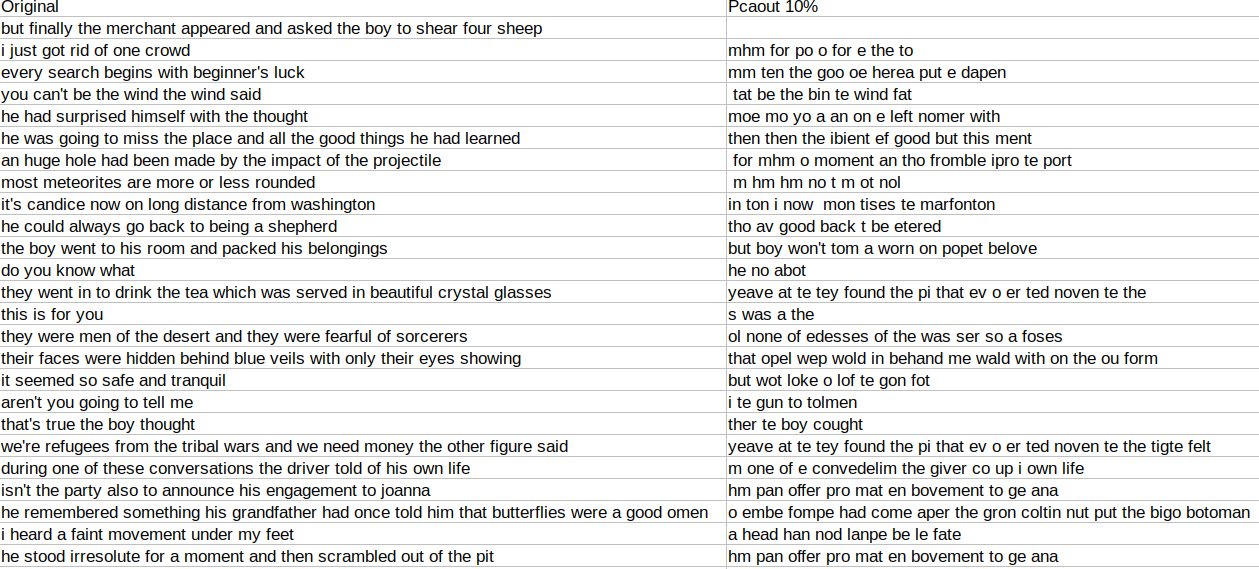}
  \caption{A snapshot of the transcribed outputs given by DeepSpeech when we only kept 10\% of the frequencies. DeepSpeech transcribes the adversarial inputs very poorly.}~\label{fig:figure8}
\end{figure}

Figure \ref{fig:figure8} shows a list of DeepSpeech outputs for the adversarial inputs crafted with PCA 10\% reconstruction to make it easier for the reader to comprehend our work. Output texts by DeepSpeech are very poor. This shows that our black-box attacks were successful against DeepSpeech when we used PCA. 
 
\begin{figure}
\centering
  \includegraphics[width=0.9\columnwidth]{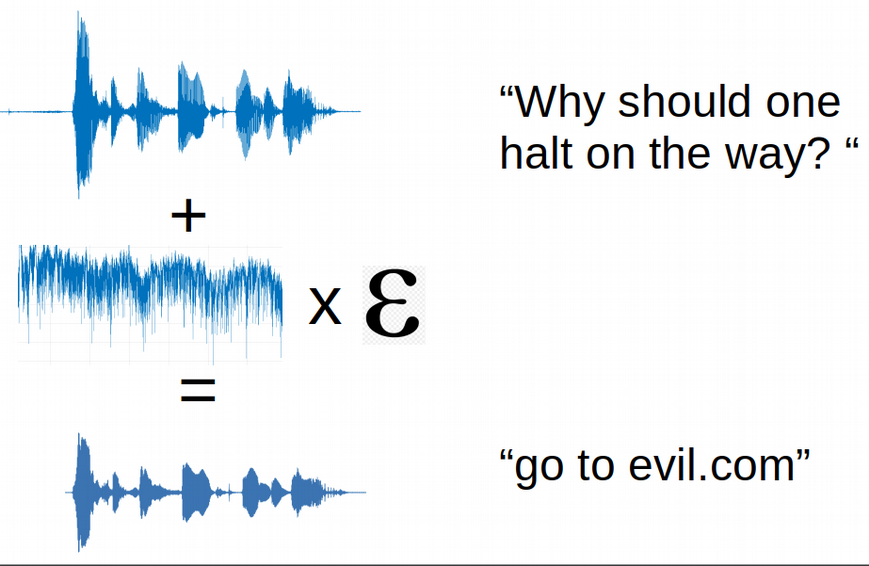}
  \caption{White-Box attacks are targeted. They are stronger but they have to have full adversarial knowledge. We can target any sentence given any arbitrary waveform with this approach. This also includes embedding voice in music, or silencing voice in a waveform. . Adversarial noise is added to the original input. In this example, the input file says "why should one halt on the way?". The targeted text is "go to evil.com". }~\label{fig:figure9}
\end{figure}

\begin{figure}
\centering
  \includegraphics[width=0.9\columnwidth]{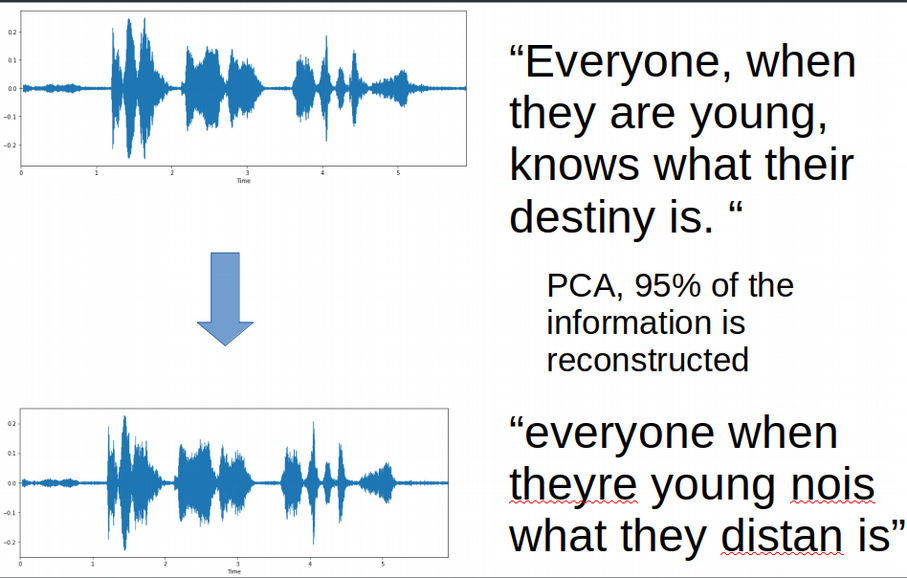}
  \caption{Black-box attacks are not targeted. They are weaker, but they don't have to have full adversarial knowledge. The input file says, "why should one halt on the way?". There is no targeted text. We apply PCA so that reconstructed adversarial input has 95\% of the original frequency components. DeepSpeech's output string has a Normalized Edit Distance of 9 to the original text. The original audio and the adversarial audio are of 87.85\% similarity. DeepSpeech fails to fully transcribe this adversarial attack. }~\label{fig:figure10}
\end{figure}
 
\begin{figure}
\centering
  \includegraphics[width=0.9\columnwidth]{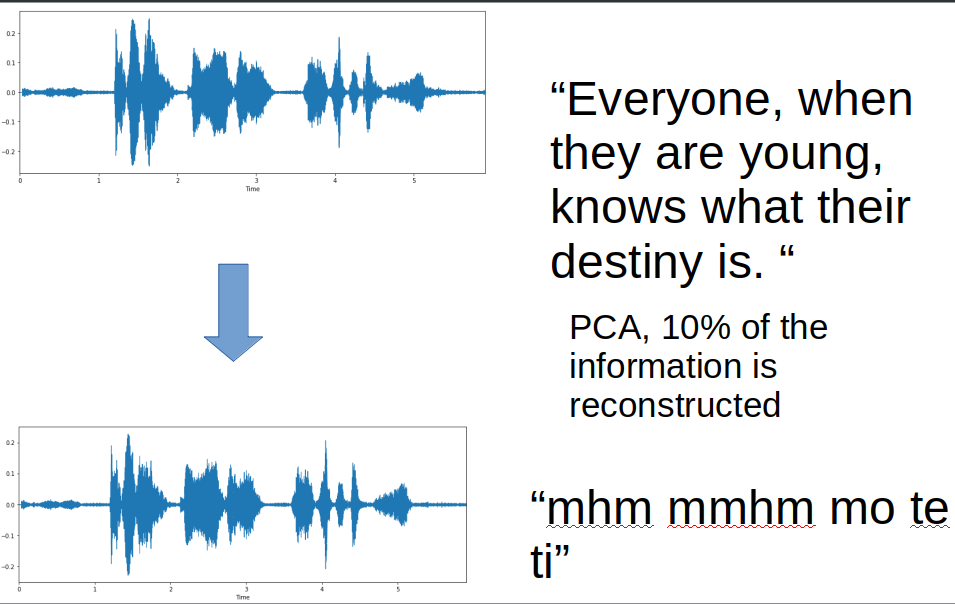}
  \caption{In this example,the input file says "why should one halt on the way?". There is no targeted text. We apply PCA so that reconstructed adversarial input only has 10\% of the original frequency components. DeepSpeech's output string has a very poor Normalized Edit Distance of 46 to the original text. Original audio and the adversarial audio are of 27.39\% similarity. DeepSpeech fails to transcribe this adversarial attack.  }~\label{fig:figure11}
\end{figure}
 
Table \ref{tab:table1} shows the percentages of frequency components that are used to reconstruct original audio signal. High percentage means most of the frequency components were kept, and reconstruction is almost perfect. Low percentage means only few of the frequency components were kept, and we lose most of the original information. Information in this context is the frequency components that make up the signal. Since the speech duration and the number of frequency components in an input are not fixed, we can \emph{NOT} hard code a number for the frequency components to be used during reconstruction(i.e we can not say 'pick 3 most primary components, 4 most primary components etc.') because we don't know the number of frequency components during speech. Speech is a continuous signal unlike images, which are a discrete signal. Instead of fixing a certain number for the primary components during PCA, we experiment with different percentage of information retain compared to the original signal, with n primary components, n being calculated based on the percentage given the input signal. (ratio of frequency components in the audio used to reconstruct the original versus all frequency components defines the percentage of the information retain in this context) 'Similarity' column in Table \ref{tab:table1} is an index that indicates how close two signals are as described  in Basic Approach section. 'Normalized Edit Distance' [23] is the number of steps necessary to convert one string to another, where actions are deletion, substitution, insertion. Higher number of actions means low similarity. 
\begin{table}[H]
\begin{center}
\begin{tabular}{l|l|p{35mm}}
\hline
\ Info. retained(\%) & Similarity (\%) & Norm. Edit Distance\\
\hline
10 & 27.472 & 46\\
35 & 37.472 & 33\\
65 & 42.216 & 22\\
80 & 73.932 & 14\\
95 & 87.850 & 9\\

\hline
\end{tabular}
\end{center}
\caption{All of the data is also shown in figure \ref{fig:figure12} and \ref{fig:figure13}.}~\label{tab:table1}
\end{table}
    
DeepSpeech performs significantly poorly in the '10\% Reconstruction' case with an average of 27\% similarity index. For the 95\% case, DeepSpeech doesn't perform as poorly as the 10\% case, but it still fails to fully transcribe the speech. 

One has to note that a human can easily tell that the audio signal has been altered during 10\% case . So, during a real-world adversary attack scenario, use of '95\%  Reconstruction' case would make more sense for the adversary since both attacks result in transcription failure, but 95\% is much similar to the original audio. So, it gets harder to detect it. We encourage reader to listen to adversarial attacks created during this study.\footnote{\url{https://www.dropbox.com/sh/0e9m4an694lkv6w/AACggV86yaopxKF_WVt8iTXNa?dl=0}}
    
\begin{figure}[!ht]
\centering
  \includegraphics[width=0.9\columnwidth]{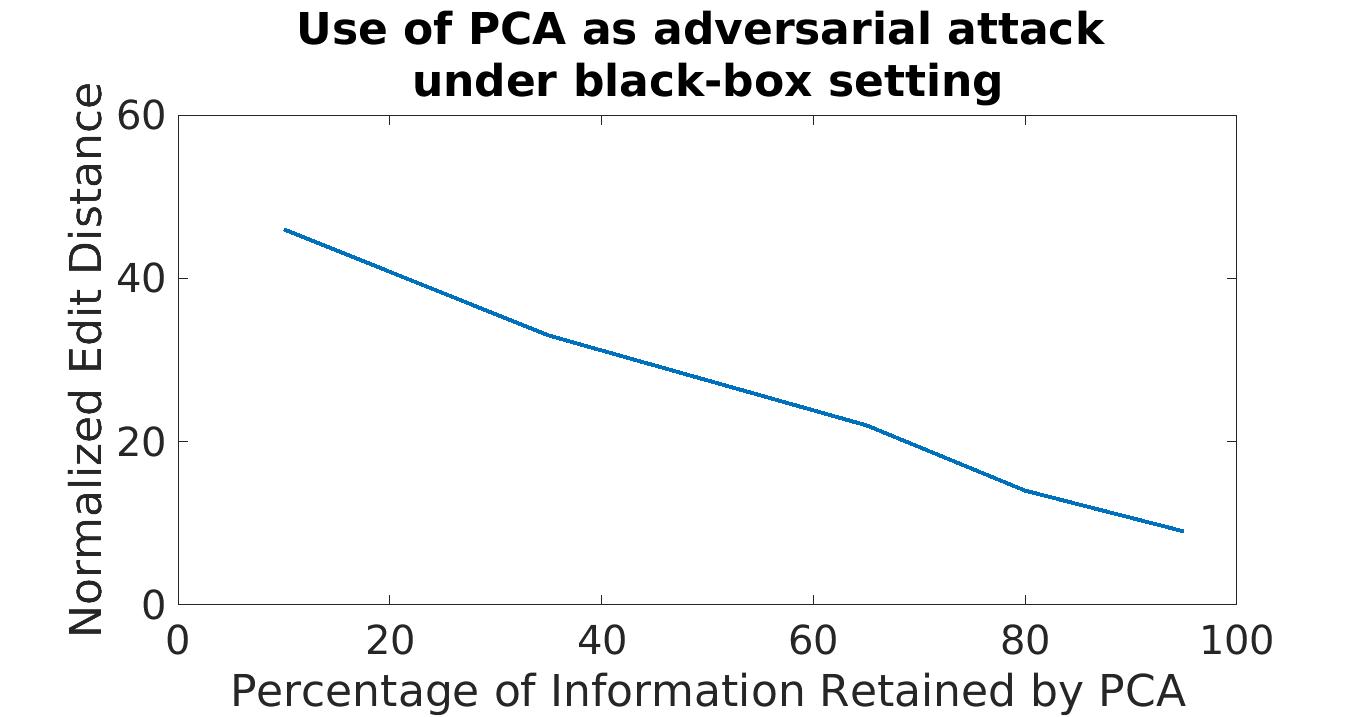}
  \caption{Average similarity distance when PCA is applied to all 25 adversarial examples with respect to percentage of components used to craft the adversarial effect during PCA.}~\label{fig:figure12}
\end{figure}

\begin{figure}[!ht]
\centering
  \includegraphics[width=0.9\columnwidth]{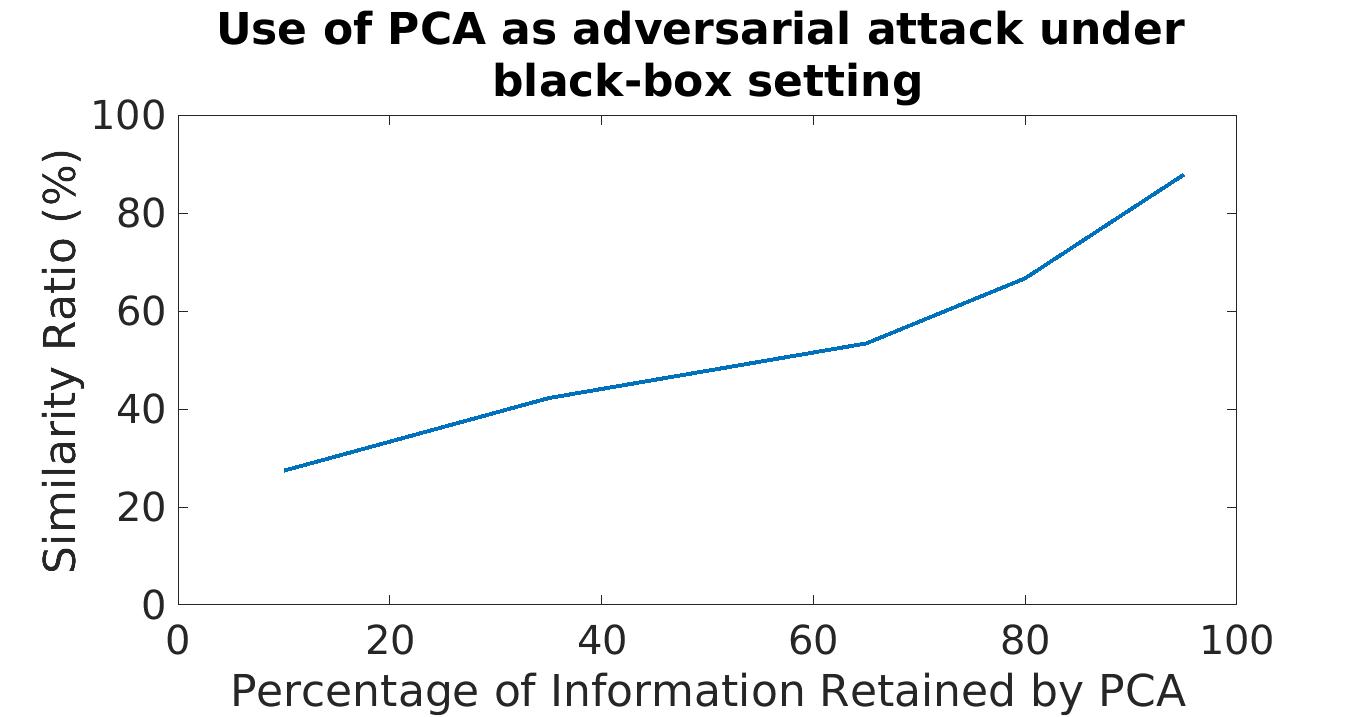}
  \caption{Average similarity distance when PCA is applied to all 25 adversarial examples with respect to percentage of components used to craft the adversarial effect during PCA}~\label{fig:figure13}
\end{figure}

Generating a single adversarial example requires approximately three-four hours of compute time on a local hardware( Intel(R) Core(TM)2 Duo CPU E8400  @ 3.00GHz 2000 with 2 CPU cores and 6144 KB cache size--No GPU ). At most, two attacks can be crafted simultaneously. Lack of GPU  is definitely a major problem whilst training Machine Learning Models.

\section{Discussion \& Concluding Remarks}
In this study, we studied applying principal components for generating adversarial attacks as well as using them to look for defense mechanisms in audio domain against adversarial attacks. We study the adversarial attacks under white-box and black-box settings.White-box targeted attacks result in 100\% adversarial success (DeepSpeech can fails to transcribe the original input and transcribes the targeted sentence instead). So, all 25 targeted attacks result in misclassification when tested against DeepSpeech. We can successfully embed voice in music. DeepSpeech transcribes Bach Cello Suite No:1 as "evil" for instance. We can successfully target silence, that is humans can hear the sentence, but DeepSpeech can't. Additionally, PCA succeeds as a way of creating black-box attacks. We were able to reconstruct 25 samples from Common Voice Data set, by keeping 95\% of the frequency components, while achieving 100\% adversarial success rate (that is, zero successful classification by DeepSpeech). When we kept 10\% of the frequencies, we also achieve 100\% adversarial success rate, but the input sentence sounds very corrupted to human ear. In the case of 95\% reconstruction, we achieve average similarity ratio of 87.85\% percent and normalized edit distance of 9, whereas in the case of 10\% reconstruction, we achieve average similarity ratio of 27.42\%, and normalized edit distance of 46. 

\section{Future Work}

Adversarial attacks in audio domain is a relatively new field. We tried to investigate white-box and black-box approached attacks and their strengths/weaknesses in audio domain. We also tried to study one defense mechanism, that is use of PCA to smooth out the adversarial noise.Future work would include to study the effect of transferability. Adversarial attacks that result in misclassification might not do so when tested against a different model. We still don't know what criteria/set of conditions dictates whether a given adversarial input will be adversary for all other models. The concept of universal adversarial attacks is still being studied [24]. 

DeepSpeech implemented by Mozilla currently serves as a benchmark in literature because of its significant accuracy (6\% word error rate over a corpus of 240 million words) [12]. Inputs that are adversary for DeepSpeech are \textit{de facto} expected to be adversary for other models. So, we are not too much worried that we couldn't complete the transferability studies although we still admit that it would be a worthwhile to complete transferability studies in the future had we had more time. 

Another point that might be worthwhile to consider is the possibility of adversarial training. Goodfellow et al showed that adversarial training can help increase accuracy [5] in image recognition domain. Such investigation would require more data, but it would be worthwhile to see the effect of altered wave forms in the training data set to achieve a more robust neural network. Overall, we hope our research may help to advance the study of adversarial attacks on RNNs and defensive mechanisms to counteract them in the audio recognition domain.

\end{document}